\documentclass[11pt,a4paper]{article}
\usepackage[hyperref]{naaclhlt2019}
\usepackage{times}
\usepackage{latexsym}
\usepackage{booktabs}
\usepackage{graphicx}
\usepackage{url}
\usepackage{array}
\usepackage{subfigure}
\usepackage{amsmath}
\usepackage{comment}
\usepackage{multirow}

\newcommand{\secref}[1]{section~\ref{#1}}
\newcommand{\Tabref}[1]{Table~\ref{#1}}
\newcommand{\tabref}[1]{table~\ref{#1}}
\newcommand{\Figref}[1]{Figure~\ref{#1}}
\newcommand{\figref}[1]{figure~\ref{#1}}

\newcommand{\appendixref}[1]{appendix~\ref{#1}}

\definecolor{roamdarkblue}{HTML}{0499CC}
\definecolor{roamlightblue}{HTML}{03A9F4}
\definecolor{roamdarkgray}{HTML}{838A8A}
\definecolor{roamlightgray}{HTML}{B8B8B8}
\definecolor{roamgreen}{HTML}{4D8951}
\definecolor{roamblack}{HTML}{212121}
\definecolor{roamsteelblue}{HTML}{9BB8D7}
\definecolor{roamorange}{HTML}{FDBA58}
\definecolor{roamwhite}{HTML}{FAFAFA}
\definecolor{roampurple}{HTML}{876DB5}

\definecolor{superlightgray}{HTML}{DDDDDD}
\definecolor{superlightgreen}{HTML}{B4FFB4}
\definecolor{superlightorange}{HTML}{FFD090}

\newcolumntype{T}[1]{>{\raggedright\arraybackslash\hspace{0pt}}p{#1}}

\newcommand{\ourlabel}[1]{\textsc{#1}}
\newcommand{\prevents}{\ourlabel{Prevents}}
\newcommand{\treats}{\ourlabel{Treats}}
\newcommand{\treatsoutcomes}{\ourlabel{Treats Outcomes}}
\newcommand{\notestablished}{\ourlabel{Not Established}}
\newcommand{\notrecommended}{\ourlabel{Not Recommended}}

\newcommand{\drug}[1][i]{\mathit{c}_{#1}}
\newcommand{\disease}[1][j]{\mathit{d}_{#1}}

\newcommand{\treatpred}{\textbf{treats}}
\newcommand{\crfpred}{\textbf{CRF\_treats}}

\newcommand{\diseaserel}{\textbf{R}}
\newcommand{\drugrel}{\hspace{1.2pt}\textbf{S}\hspace{1.2pt}}
\newcommand{\pos}{{\color{white} \neg}}

\aclfinalcopy 

\title{Modeling Drug--Disease Relations \\ with Linguistic and Knowledge Graph Constraints}

\author{
  Bruno Godefroy\\
  Roam Analytics\\
  \And
  Christopher Potts\\
  Roam Analytics\\
  Stanford University\\
}

\date{}

\begin{document}

\maketitle


\begin{abstract}
  FDA drug labels are rich sources of information about drugs and
  drug--disease relations, but their complexity makes them challenging
  texts to analyze in isolation. To overcome this, we situate these
  labels in two health knowledge graphs: one built from precise
  structured information about drugs and diseases, and another built
  entirely from a database of clinical narrative texts using simple
  heuristic methods. We show that Probabilistic Soft Logic models
  defined over these graphs are superior to text-only and
  relation-only variants, and that the clinical narratives graph
  delivers exceptional results with little manual effort. Finally, we
  release a new dataset of drug labels with annotations for five
  distinct drug--disease relations.
\end{abstract}


\section{Introduction}\label{sec:introduction}

The FDA Online Label Repository is a publicly available database of
texts that provide detailed information about pharmaceutical drugs,
including active ingredients, approved usage, warnings, and
contraindications. Although the labels have predictable subsections
and use highly regulated language, they are nuanced and presuppose
deep medical knowledge, since they are intended for use by healthcare
professionals \citep{shrank2007educating}. This makes them a
formidable challenge for information extraction systems.

\begin{figure}[t]
  \centering
  \includegraphics[width=1\linewidth]{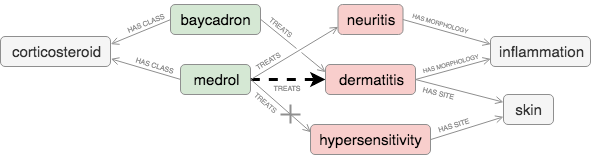}
  \caption{Illustrative knowledge graph for the prediction of the
    relation between a drug (Medrol) and a disease (dermatitis). In
    favor of a \treats\ relation: Medrol has the same pharmacologic
    class as Baycadron, which treats dermatitis; dermatitis has the
    same morphology as neuritis, which is treated by Medrol. Against a
    \treats\ relation: dermatitis has the same site as
    hypersensitivity, which is not treated by Medrol. Our
    Probabilistic Soft Logic approach learns to combine these
    factors.%
  }
  \label{fig:toy_example}
\end{figure}

At the same time, existing medical ontologies contain diverse
structured information about drugs and diseases. The drug labels can
be situated in a larger health knowledge graph that brings together
these sources of information, which can then be used to understand the
labels. \Figref{fig:toy_example} illustrates the guiding intuition; if
our goal is to determine whether to add the dashed \treats\ edge, we
should take advantage of the label for Medrol as well as the
drug--drug, drug--disease, and disease--disease relations that are
observed. \citet{hristovski2006exploiting} describe a similar intuition
with what they call ``discovery patterns".

In this paper, we show that Probabilistic Soft Logic (PSL;
\citealt{Brocheler-etal:2010,bach2013hinge}) is an effective tool for
modeling drug labels situated in larger health knowledge graphs.  A
PSL model is a set of logical statements interpreted as defeasible
relational constraints, and the relative weights of these constraints
are learned from data. In PSL, we can directly state intuitions like
`if some drugs share a substance, then they might treat
the same diseases' and `if two diseases share a classification,
they might be treated by the same drugs'. \citet{fakhraei2013drug} show the
value of these ideas for drug--disease relationships. We
extend their basic idea by adapting the method of
\citet{West-etal:2014}: our PSL models combine graph-based constraints
with constraints derived from a separate sequence-labeling model
applied to the drug label texts.

The most expensive parts of this model are the structured ontologies
used to build the graph. Each ontology is a large, highly specialized
project requiring extensive human expertise. This motivates us to seek
out alternatives. To this end, we also evaluate on a graph that is
derived entirely from clinical texts. Using only key-phrase matching,
we instantiate a large, noisy graph with the same core structure as in
\figref{fig:toy_example}, and we show that the PSL model is robust to
such noisy inputs. By combining this graph with our structured one, we
achieve larger gains, but the clinical narratives graph is effective
on its own, and thus emerges as a viable option for domains lacking
pre-built ontologies.

Our evaluations are centered around a new dataset of annotated drug
labels. In this dataset, spans of text identifying diseases are
annotated with labels summarizing the drug--disease relationship
expressed in the
label.\footnote{\url{https://github.com/roamanalytics/roamresearch/tree/master/Papers/PSL-drugs-diseases}}
We show that our full PSL model yields superior
predictions on this dataset, as compared with ablations using only the
texts and only the graph edges derived from structured and textual
sources.


\section{Medical Ontologies Knowledge Graph}\label{sec:graph}

Our health knowledge graphs are focused on drugs and diseases related
to obesity. We first introduce the graph we built from structured
sources (802 nodes, 883 edges).

\subsection{FDA Online Label Repository}

From the FDA Online Label
Repository\footnote{\url{https://labels.fda.gov}} we extracted the
drug labels matching at least one of the following keywords:
``obesity'', ``overweight'', ``asthma'', ``coronary heart disease'',
``hypercholesterolemia'', ``gallstones'', ``gastroesophageal reflux'',
``gout'', ``hypertriglyceridemia'', ``sleep apnea'', ``peripheral
vascular disease'', ``chronic venous insufficiencies''. This is the
set of diseases in the i2b2 Obesity Challenge \citep{Uzuner:2009}, with
some omissions for the sake of precision.

The FDA individuates drugs in a very fine-grained way, resulting in
many duplicate or near-duplicate labels (there are usually multiple
brands and versions of a drug). To ensure no duplicates in the
intuitive sense, we hand-filtered to a set of 106 drug
labels. These labels mention 198 distinct diseases, resulting in 1,110
drug--disease pairs.

\subsection{Drug--Disease Annotations}\label{sec:ann}

\begin{table}[!htb]
  \centering
  \setlength{\tabcolsep}{2pt}
  \begin{tabular}[c]{*{6}{r}}
    \toprule
             & Relations & Agreement \\
    \midrule
    \prevents         & 154   & 63.0 \\
    \treats           & 4,425 & 67.3 \\
    \treatsoutcomes   & 2,268 & 67.1 \\
    \notestablished   & 241   & 35.1 \\
    \notrecommended   & 262   & 49.5 \\
    \ourlabel{other}  & 1,262 & 35.9 \\
   \bottomrule
  \end{tabular}
  \caption{Drug--disease relations collected using crowdsourcing. The
    ``Agreement'' column gives the average agreement between workers
    and the labels inferred. Higher agreement correlates strongly with
    the degree to which the information is explicit in the label text.}
  \label{tab:labeldistribution}
\end{table}

Disease mentions in drug labels have a variety of senses.  Guided by
our own analysis of the relevant sentences, we settled on the
following set of relational descriptions, with input from clinical
experts acting as consultants to us:
\begin{itemize}\setlength{\itemsep}{0pt}
\item The drug prevents the disease (our label is \prevents).
\item The drug treats the disease (\treats).
\item The drug treats outcomes of the disease (\treatsoutcomes).
\item The safety/effectiveness of the drug has not been established
  for the disease (\notestablished).
\item The drug is not recommended for the disease (\notrecommended).
\end{itemize}
We emphasize the clinical importance of distinguishing between
treating a disease and treating its outcomes. Cancer, for example, is
not treated with Depo-Medrol, but hypercalcemia associated with cancer
is. Similarly, \notrecommended\ identifies a contraindication, a
specific kind of guidance, whereas the superficially similar
\notestablished\ has a more open meaning: the relation could be
\treats, but this has not been tested or a clinical trial failed to
demonstrate its efficacy.

We crowdsourced the task of assigning these labels to disease mention
spans. Workers saw the entire label text, with our target diseases
highlighted, and were asked to select the best statement among those
provided above. We launched our task on Figure Eight, asking for 5
judgments for each drug--disease pair. To infer a label for each
example from these responses, we applied Expectation Maximization
(EM), which estimates the reliability of each worker and weights their
contributions accordingly \citep{dawid1979maximum}.
\Tabref{tab:labeldistribution} reports average agreement between
workers and the inferred label. The \prevents, \treats, and
\treatsoutcomes\ relations are usually stated directly in drug labels,
leading to high agreement. In contrast, \notestablished,
\notrecommended, and \ourlabel{Other} are more subtle and diverse,
leading to lower agreement. The label of Normosol-R, for example,
states: ``The solution is not intended to supplant transfusion of
whole blood or packed red cells in the presence of uncontrolled
hemorrhage''.  What is the relation between Normosol-R and hemorrhage?
In this case, any of \notestablished, \notrecommended, and
\ourlabel{Other} seems acceptable.

We observe a correlation between the number of distinct disease
mentions in a drug label and its ratio of \treats\ labels against the
other labels. Some drugs are more ``general purpose'' than others, in
that they are involved in many treatment relations. For example, the
label for Prednisone contains mentions of 50 distinct diseases it can
treat.  To prevent our system from predicting treatment relations
primarily using the out-degree of drug nodes (instead of the domain
knowledge provided by the graph), we manually removed 19 of these
high-degree drugs.  After removing these drugs, our dataset contained
431 relations.

\subsection{Structured Medical Ontologies}\label{sec:ontologies}

Our graph contains structured information about drugs and diseases from
a variety of sources.
Our observations for drugs are route
(OpenFDA),\footnote{\url{https://open.fda.gov}} pharmacologic class
(OpenFDA), substances
(OpenFDA),
and dosage form (RxNorm).\footnote{\url{https://www.nlm.nih.gov/research/umls/rxnorm/}}
Our observations for diseases are all from
SNOMED~CT:\footnote{\url{https://www.snomed.org}} %
finding sites, %
associated morphologies, %
and courses.

In general, these ontologies are quite sparse; a missing edge is
therefore not necessarily evidence for a missing relationship. Broadly
speaking, this is why text analysis can be so meaningful in this
context -- structured resources always fall behind because of the
challenges of manual creation. This sparsity also motivates the
approximate, text-based graph we introduce next.


\section{Clinical Narratives Knowledge Graph}\label{sec:text-graph}

The Medical Ontologies Knowledge Graph is precise, but the underlying
resources are expensive to build, which creates sparsity and can be an
obstacle to adapting the ideas to new domains. To try to address this,
we built a comparable graph using only de-identified clinical
narratives -- reports (often transcribed voice recordings) of
clinicians' interactions with patients.

Using lexicon-based matching methods, we extracted drug and disease
mentions from these texts. In the resulting graph, each text is a
node, with edges to the drug and disease nodes corresponding to these
extracted mentions. The resulting graph has 319,598 nodes and 421,502
edges.

This is a simpler graph structure than we are able
to obtain from structured resources (cf.~\secref{sec:ontologies}), but
it supports our most important logical connections. For example, we
can say that if diseases $\disease[1]$ and $\disease[2]$ are mentioned
in the same narrative and $\disease[1]$ is treated by a specific drug,
then $\disease[2]$ might also be treated by that drug. These hypotheses are
often not supported; for example, a patient with a number of unrelated
medical conditions might lead to many false instances of this
claim. Nonetheless, we expect that, in aggregate, these
connections will prove informative.


\section{Models}\label{sec:model}

\begin{table*}[htp]
  \centering
  \renewcommand{\arraystretch}{1.2}
  $\begin{array}[c]{@{}
     r @{\hspace{14pt}}
     c @{\hspace{24pt}}
     l @{\thickspace}
     @{}}

   \multirow{2}{*}{Priors}
   &
   & \pos\treatpred(\drug, \disease)
   \\
   &
   & \neg\treatpred(\drug, \disease)
   \\[2ex]

   \text{1a.}
   & \multirow{2}{*}{\includegraphics[height=32pt]{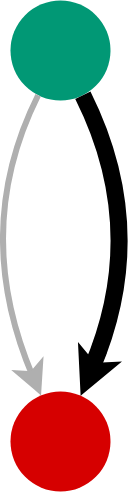}}
   & \pos\crfpred(\drug[i], \disease[j])
     \thickspace \Rightarrow \thickspace
     \pos\treatpred(\drug[i], \disease[j])
   \\
   \text{b.}
   &
   & \neg\crfpred(\drug[i], \disease[j])
     \thickspace \Rightarrow \thickspace
     \neg\treatpred(\drug[i], \disease[j])
   \\[2ex]

   \text{2a.}
   & \multirow{2}{*}{\includegraphics[height=32pt]{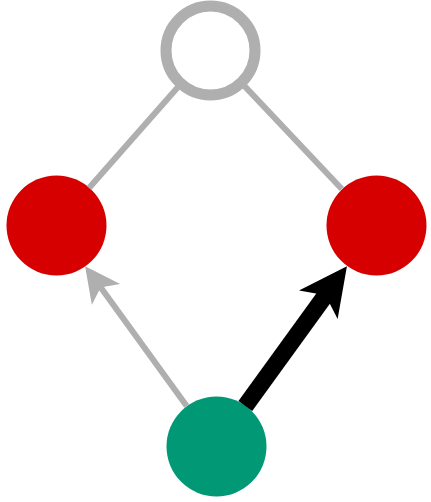}}
   & \diseaserel(\disease[1], x)
     \thickspace\wedge\thickspace \diseaserel(\disease[2], x)
     \thickspace\wedge\thickspace \disease[1] \neq \disease[2]
     \thickspace\wedge\thickspace \pos\treatpred(\drug[i], \disease[1])
     \thickspace\Rightarrow\thickspace \pos\treatpred(\drug[i], \disease[2])
   \\
   \text{b.}
   &
   & \diseaserel(\disease[1], x)
     \thickspace\wedge\thickspace \diseaserel(\disease[2], x)
     \thickspace\wedge\thickspace \disease[1] \neq \disease[2]
     \thickspace\wedge\thickspace \neg\treatpred(\drug[i], \disease[1])
     \thickspace\Rightarrow\thickspace \neg\treatpred(\drug[i], \disease[2])
   \\[2ex]

   \text{3a.}
   & \multirow{2}{*}{\includegraphics[height=32pt]{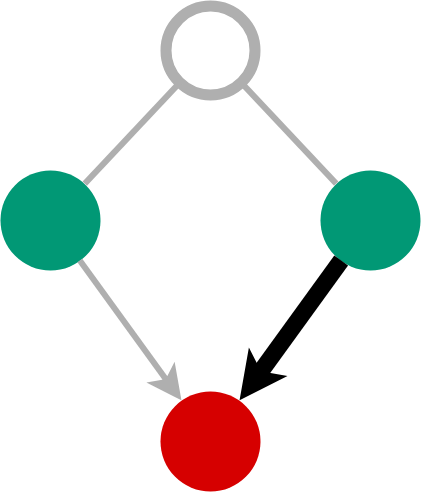}}
   & \drugrel(\drug[1], x)
     \thickspace\wedge\thickspace \drugrel(\drug[2], x)
     \thickspace\wedge\thickspace \drug[1] \neq \drug[2]
     \thickspace\wedge\thickspace \pos\treatpred(\drug[1], \disease[j])
     \thickspace\Rightarrow\thickspace \pos\treatpred(\drug[2], \disease[j])
   \\
   \text{b.}
   &
   & \drugrel(\drug[1], x)
     \thickspace\wedge\thickspace \drugrel(\drug[2], x)
     \thickspace\wedge\thickspace \drug[1] \neq \drug[2]
     \thickspace\wedge\thickspace \neg\treatpred(\drug[1], \disease[j])
     \thickspace\Rightarrow\thickspace \neg\treatpred(\drug[2], \disease[j])
  \end{array}$
  \caption{PSL rules. The diagrams show inferred edges (black) based on
    observed nodes and edges (gray), drug nodes (green), and disease nodes (red).
    Variables $\drug[i]$ and
    $\disease[j]$ range over drugs and diseases, respectively.  The first
    two rules serve as priors on the
    $\treats$ relation.  The relation
    $\diseaserel$ can be \textbf{has\_associated\_morphology},
    \textbf{has\_course}, or \textbf{has\_finding\_site}.  The
    relation
    $\drugrel$ can be \textbf{has\_route}, \textbf{has\_substance},
    \textbf{has\_doseform}, or \textbf{has\_pharmclass}. The variable
    $x$ in turn ranges over the semantically appropriate entities
    given the relation in question.  For the Clinical Narratives Graph, both
    $\diseaserel$ and
    $\drugrel$ have only the value \textbf{is\_mentioned\_in} and
    $x$ ranges over clinical narratives.}
  \label{tab:rules}
\end{table*}

Our core task is to identify drug--disease \treats\ relations based on
the drug's label text. We consider text-only, graph-only, and combined
approaches to this problem.

\subsection{Text-Only Model}

Our text-only model is a separately optimized conditional random
fields (CRF; \citealt{Lafferty:McCallum:Pereira:2001}) model trained
on 2,000 annotated sentences sampled from the full dataset of FDA drug
labels. When this CRF is used in isolation, we say that a drug
$\drug[i]$ treats a disease $\disease[j]$ just in case our trained CRF
identifies at least one \treats\ span describing $\disease[j]$ in the
label for $\drug[i]$. To incorporate this CRF into our PSL models, we
simply add two logical statements: one to influence the prediction
confidence positively, another negatively (\tabref{tab:rules}, rules
1a,b). This encodes the goal of agreeing with this model's
predictions.

\subsection{PSL Graph Rules}

Our PSL rules fall into a few major classes. The guiding intuitions
are that (i) relations connecting the same drug--disease pairs should
have the same confidence, (ii) drugs that have nodes in common should
treat the same diseases, and (iii) diseases that have nodes in common
should be treated by the same drugs. \Tabref{tab:rules} schematizes
the full set of rules that we define over both our graphs.

\subsection{Optimization}

We use the Probabilistic Soft Logic
package.\footnote{\url{http://psl.linqs.org}} Details about our
optimization choices are given in \appendixref{app:optimization}. The rules
for our full model (all instances of the schemas in
\tabref{tab:rules}), along with their learned weights and groundings
(supporting graph configurations), are given in
\appendixref{app:weights}.


\section{Experiments}\label{sec:experiments}

\Figref{fig:results} reports our main results (see
appendix~\ref{app:experimental-details} for additional experimental
details). Both the CRF and the graph rules make positive
contributions, and any combination of these two sources of information
is superior to its text-only or graph-only ablations. Furthermore, our
Clinical Narratives Graph proves extremely powerful despite its
simplistic construction. Numerically, our best model for essentially
any amount of evidence is one that includes rules that reason about
both graphs in addition to the CRF predictions, though using the
Narratives Graph alone is highly competitive with this larger
one. These findings strongly support our core hypothesis that
information extraction from drug labels can benefit from both text
analysis and graph-derived
information.

\begin{figure}[t]
  \centering
  \includegraphics[width=0.9\linewidth]{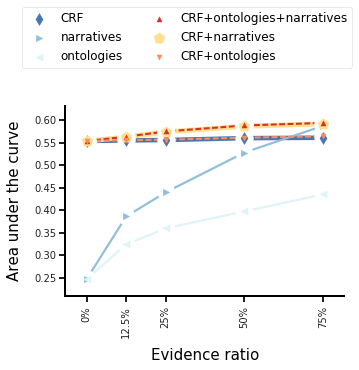}
  \vspace{-2mm}
  \caption{Area under the Precision-Recall curve (AUC) as function of
    evidence ratio (proportion of drug-disease edges that are observed).
    We favor AUC over receiver--operator curve (ROC) because
    the dataset is highly imbalanced -- only 24.8\% of relations are
    positive for the \treats\ label. Results are averaged over 100
    runs for each model.}
  \label{fig:results}
\end{figure}


\section{Conclusion}\label{sec:conclusion}

We presented evidence that combining text analysis with PSL rules
defined over a knowledge graph can improve predictions about
drug--disease treatment relations, and we showed that this can be
effective even where the graph is derived heuristically from clinical
texts, which means that the techniques can be applied even in domains
that lack rich, broad coverage ontologies.


\bibliography{psl-paper-bib}
\bibliographystyle{acl_natbib}

\clearpage

\newpage

\appendix

\section{Supplemental Material}

\setcounter{table}{0}
\renewcommand{\thetable}{A\arabic{table}}

\subsection{Experimental Details}\label{app:experimental-details}

For each run, we sample two distinct sets of diseases. Then, with each
set, we create a subgraph by considering all the drugs adjacent to
these diseases, and the edges between these drugs and diseases. The
two subgraphs have distinct disease nodes, distinct edges, and some
drug nodes in common.  One of these subgraphs is used for training,
the other one for evaluation.  In each subgraph, we sample 25\% of the
edges that are used for prediction. Depending on the evidence ratio,
some of the remaining 75\% of the edges are provided as observations
to the model. For evidence ratio = 0.75 for example, 75\% of the
edges in each subgraph are observed and the other 25\% are predicted
by models.

\subsection{Optimization details}\label{app:optimization}

In training, initial weight values are very important. Whatever the
learning rate and number of steps, it is easy to get stuck in local
optima. We computed the initial weights using the number of groundings
for each rule, such that (i) each source of information (CRF,
ontologies, narratives) has the same contribution, and (ii) for a
given source of information, each rule has the same contribution.

The weights were learned by optimizing the pseudo-log-likelihood of
the data using the voted perceptron algorithm, as implemented by the
PSL package. We preferred the pseudo-log-likelihood to the
log-likelihood for scalability reasons.

Each model was trained over 10 iterations, with a training step of 1.

For inference, we used the Alternating Direction Method of Multipliers
(ADMM; \citealt{boyd2011distributed}), as implemented by the PSL
package. Consensus and local variables were initialized to a fixed
value (0.25, close to the true positive \treats\ edge ratio), instead
of randomly, to speed up convergence. The absolute and relative error
components of stopping criteria were set to $10^{-6}$, with a maximum
of 25,000 iterations.

\subsection{Learned weights and groundings}\label{app:weights}

\newcommand{\positiverule}{a}
\newcommand{\negativerule}{b}

\begin{table*}[h]
  \centering
  \begin{tabular}[c]{l r r}
    \toprule
    Rule & Relative learned weight & Groundings \\
    \midrule
    Prior (positive) & 1.10 & 431 \\
    Prior (negative) & 0.87 & 431 \\
    Rule 1\positiverule\ & 1.02 & 107 \\
    Rule 1\negativerule\ & 0.98 & 324 \\
    Rule 2\positiverule\ with $\diseaserel = \textbf{has\_associated\_morphology}$ & 1.05 & 3 \\
    Rule 2\negativerule\ with $\diseaserel = \textbf{has\_associated\_morphology}$ & 1.11 & 30 \\
    Rule 2\positiverule\ with $\diseaserel = \textbf{has\_course}$ & $-$    & 0 \\
    Rule 2\negativerule\ with $\diseaserel = \textbf{has\_course}$ & 0.97 & 14 \\
    Rule 2\positiverule\ with $\diseaserel = \textbf{has\_finding\_site}$ & 1.13 & 37 \\
    Rule 2\negativerule\ with $\diseaserel = \textbf{has\_finding\_site}$ & 0.88 & 45 \\
    Rule 3\positiverule\ with $\drugrel = \textbf{has\_route}$ & 3.09 & 347 \\
    Rule 3\negativerule\ with $\drugrel = \textbf{has\_route}$ & 0.00 & 1,095 \\
    Rule 3\positiverule\ with $\drugrel = \textbf{has\_substance}$ & 1.36 & 65 \\
    Rule 3\negativerule\ with $\drugrel = \textbf{has\_substance}$ & 1.00 & 140 \\
    Rule 3\positiverule\ with $\drugrel = \textbf{has\_doseform}$ & 1.99 & 164 \\
    Rule 3\negativerule\ with $\drugrel = \textbf{has\_doseform}$ & 0.77 & 384 \\
    Rule 3\positiverule\ with $\drugrel = \textbf{has\_pharmclass}$ & 1.00 & 52 \\
    Rule 3\negativerule\ with $\drugrel = \textbf{has\_pharmclass}$ & 1.00 & 101 \\
    Rule 2\positiverule\ with $\diseaserel = \textbf{is\_mentioned\_in}$ & 22.12 & 29,803 \\
    Rule 2\negativerule\ with $\diseaserel = \textbf{is\_mentioned\_in}$ & 0.0 & 115,251 \\
    Rule 3\positiverule\ with $\drugrel = \textbf{is\_mentioned\_in}$ & 7.42 & 2,568 \\
    Rule 3\negativerule\ with $\drugrel = \textbf{is\_mentioned\_in}$ & 0.90 & 14,062 \\
    \bottomrule
  \end{tabular}
  \caption{Learned weights and groundings in our full model (evidence
    ratio = 0.75). Rule numbers refer to the schemas in
    \tabref{tab:rules}. Weights are relative to their initial value
    (ratio of the learned weight over its respective initial weight).
    Weights assigned to ``positive'' rules are often larger than the
    weight of their respective negative rule, which suggests that
    positive groundings are usually more informative than negative
    ones. For example, drugs that have the same route are more likely
    to have the same positive \treats\ edges than to have the same
    negative \treats\ edges.
  }
\end{table*}

\end{document}